\documentclass[a4paper,fleqn]{cas-dc}

\usepackage[numbers]{natbib}
\usepackage{lineno,hyperref}
\usepackage{booktabs}
\usepackage{bookmark}
\usepackage{geometry}
\usepackage{tabularx}
\usepackage{graphicx}
\usepackage{multirow}
\usepackage{colortbl}
\usepackage{url}
\usepackage{bm}
\usepackage{amssymb,amsmath}
\usepackage{amsthm}
\usepackage{forest}
\usepackage{xspace}
\usepackage{CJK}
\usepackage[caption=false,font=footnotesize]{subfig}
\usepackage[ruled,linesnumbered]{algorithm2e}
\usepackage{xparse}
\usepackage{appendix}
\usepackage{subcaption} 

\newtheorem{definition}{Definition}

\def\tsc#1{\csdef{#1}{\textsc{\lowercase{#1}}\xspace}}
\tsc{WGM}
\tsc{QE}
\tsc{EP}
\tsc{PMS}
\tsc{BEC}
\tsc{DE}

\usepackage{amsmath,amsfonts,bm}

\def\eqref#1{equation~\ref{#1}}

\def\1{\bm{1}}

\DeclareMathAlphabet{\mathsfit}{\encodingdefault}{\sfdefault}{m}{sl}
\SetMathAlphabet{\mathsfit}{bold}{\encodingdefault}{\sfdefault}{bx}{n}

\def\sX{{\mathbb{X}}}

\usepackage{array}
\newcolumntype{L}[1]{>{\raggedright\let\newline\\\arraybackslash\hspace{0pt}}m{#1}}
\usepackage{paralist}

\usepackage{hyperref}
\usepackage{url}

\usepackage{array}
\newcolumntype{L}[1]{>{\raggedright\let\newline\\\arraybackslash\hspace{0pt}}m{#1}}
\usepackage{paralist}

\usepackage{hyperref}
\usepackage{url}

\begin{document}
\let\WriteBookmarks\relax
\def\floatpagepagefraction{1}
\def\textpagefraction{.001}
\let\printorcid\relax 

\shorttitle{}

\shortauthors{Huang et~al.}

\title [mode = title]{Evidential Deep Partial Multi-View Classification With Discount Fusion}                      



%
\author[a]{Haojian Huang}
\credit{Methodology, Formal analysis, Software, Validation, Visualization, Data curation, Writing - original draft}

\author[b]{Zhe Liu\corref{cor1}}
\credit{Conceptualization, Methodology, Formal analysis, Investigation, Writing - original draft, Writing - review \& editing}

\author[b]{Sukumar Letchmunan}
\credit{Investigation, Validation, Visualization, Writing - review \& editing}

\author[c,d,e]{Muhammet Deveci}
\credit{Conceptualization, Formal analysis, Writing - review \& editing, Supervision}

\author[f]{Mingwei Lin}
\credit{Formal analysis, Validation, Writing - review \& editing}

\author[g]{Weizhong Wang}
\credit{Validation, Writing - review \& editing}

\affiliation[a]{organization={Department of Computer Science, The University of Hong Kong},
    city={Hong Kong},
    postcode={999077}, 
    country={China}}

\affiliation[b]{organization={School of Computer Sciences, Universiti Sains Malaysia},
    city={Penang},
    postcode={11800},
    country={Malaysia}}

\affiliation[c]{organization={Department of Industrial Engineering, Turkish Naval Academy, National Defence University},
    city={Istanbul},
    postcode={34942},
    country={Turkey}}

\affiliation[d]{organization={Royal School of Mines, Imperial College London},
    city={London},
    postcode={SW7 2AZ}, 
    country={UK}}

\affiliation[e]{organization={Department of Information Technologies, Western Caspian University},
    city={Baku},
    postcode={1001},
    country={Azerbaijan}}

\affiliation[f]{organization={College of Computer and Cyber Security, Fujian Normal University},
    city={Fuzhou},
    postcode={350117}, 
    country={China}}

\affiliation[g]{organization={School of Economics and Management, Anhui Normal University},
    city={Wuhu},
    postcode={241000},
    country={China}}




\begin{abstract}
Incomplete multi-view data classification poses significant challenges due to the common issue of missing views in real-world scenarios. Despite advancements, existing methods often fail to provide reliable predictions, largely due to the uncertainty of missing views and the inconsistent quality of imputed data. To tackle these problems, we propose a novel framework called Evidential Deep Partial Multi-View Classification (EDP-MVC). Initially, we use K-means imputation to address missing views, creating a complete set of multi-view data. However, the potential conflicts and uncertainties within this imputed data can affect the reliability of downstream inferences. To manage this, we introduce a Conflict-Aware Evidential Fusion Network (CAEFN), which dynamically adjusts based on the reliability of the evidence, ensuring trustworthy discount fusion and producing reliable inference outcomes. Comprehensive experiments on various benchmark datasets reveal EDP-MVC not only matches but often surpasses the performance of state-of-the-art methods.
\end{abstract}


\begin{highlights}
\item A conflict-aware fusion network with learnable discount factors is proposed for dynamic multi-view evidence fusion.
\item An uncertainty-informed learning paradigm is introduces to quantify and manage uncertainties.
\item The proposed method achieves superior results on multiple real datasets, outperforming state-of-the-art methods.
\item The proposed method excels on datasets with conflicting and incomplete multi-view data across various missing rates.
\end{highlights}


\maketitle

\section{Introduction}
\label{sec:intro}
Multi-view learning harnesses diverse data sources to uncover deeper, more nuanced patterns, significantly outperforming single-view learning in understanding complex tasks. This approach not only enhances performance in applications like clustering~\cite{liu2023adaptive,zhao2023mvcformer,liu2024adaptive} and classification~\cite{han2022trusted,liu2023effective,xu2024ensembling} but also excels in multi-modal research~\cite{huang2024crest,chen2024finecliper}. By integrating deep learning, multi-view learning further amplifies its strengths, clearly demonstrating its superiority in advanced data analysis and proving indispensable for insightful outcomes.

However, in real-world scenarios, data collection and preservation issues often result in incomplete datasets~\cite{liu2024enhanced}. For instance, archival data may degrade or be lost over time. In the medical field, patients with the same condition might undergo different examinations, leading to incomplete or misaligned multi-view data. Similarly, automotive sensors might occasionally malfunction, capturing only partial information. These missing views introduce high uncertainty, making the flexible and reliable utilization of incomplete multi-view data a formidable challenge~\cite{liu2024representing}. 

Existing research on incomplete multi-view classification (IMVC) typically falls into two categories: methods that classify using only the available views without imputation, and deep learning-based generative methods, such as autoencoder~\cite{AE}, generative adversarial network (GAN)~\cite{GAN}, VAE~\cite{MVAE}, which reconstruct missing data before classification. Both approaches have notable limitations. Methods relying solely on available views often perform poorly under high missing rates due to their inability to explore inter-view correlations, leading to incomplete utilization of the multi-view data's potential. On the other hand, deep learning-based imputation methods, while innovative, frequently suffer from a lack of interpretability. These methods do not explicitly measure the importance and uncertainty of the imputed latent views, making the results less reliable. Moreover, these methods often struggle with complex missing patterns, particularly in datasets with more than two views. Even advanced methods like UIMC~\cite{xie2023exploring}, which are designed to handle such data, often neglect the potential conflicts between views. When employing the Dempster-Shafer theory framework for multi-view evidence integration, these conflicts can result in performance degradation and counterintuitive outcomes~\cite{huang2023belief,liu2023new,xu2024reliable,liu2024effective}. This highlights the need for a more robust approach that can effectively address the uncertainties and conflicts inherent in incomplete multi-view data.

To overcome these limitations, we propose a simple yet effective conflict-aware classification model for incomplete multi-view classification. It begins by imputing each missing view using a K-means-based method and gathers evidence from the imputed multi-view data, explicitly measuring potential uncertainties. Recognizing that evidence from different views may conflict, we introduce an uncertainty-driven adaptive learning conflict-robust evidence fusion network. This network employs learnable discount factors to mitigate conflicts during the evidence fusion process, ensuring more reliable and collaborative decision-making.

The main contributions of this work are summarized as follows:
\begin{itemize}
    \item We identify the sensitivity of the Dempster-Shafer theory framework to evidence conflicts during the multi-view evidence fusion process. To mitigate this issue, we introduce the Conflict-Aware Evidential Fusion Network (CAEFN). To the best of our knowledge, this is the first method to utilize uncertainty and demonstrate robustness to conflicts in incomplete multi-view classification.
    \item To address the potential uncertainties and conflicts in incomplete data, we propose an explicit reliability measurement for the evidence from each view. This reliability measurement serves as a learnable discount factor, allowing for the dynamic and adaptive fusion of evidence from multiple views.
    \item We conduct extensive quantitative and adversarial experiments on multiple incomplete multi-view benchmarks. The results show that our proposed method achieves competitive or superior performance compared to state-of-the-art methods, both on regular data and on datasets with conflicts and noise. 
\end{itemize}

The remainder of this article is organized as follows. Section~\ref{sec:Related Work} provides an overview of related work in incomplete multi-view classification and uncertainty-informed deep learning. Section~\ref{sec:method} details the proposed EDP-MVC method. In Section~\ref{sec:Experiments}, the proposed method is evaluated using seven existing benchmarks and adversarial experiments to comprehensively assess its performance and robustness. Section~\ref{sec:conclusion} concludes the paper and highlights future research directions.

\section{Related Work}
\label{sec:Related Work}
\subsection{Incomplete Multi-View Learning}
In the context of incomplete multi-view learning, handling missing views typically follows two primary strategies: ignoring the missing views or reconstructing them using sophisticated deep learning techniques. \textbf{Non-Imputation Strategies.} Some methods choose to utilize only the existing views, directly learning a common latent subspace or representation applicable to all views. This approach is widely used in clustering tasks such as MCP, PMC, and IMVSG~\cite{MCP, PMC, IMVSG}, and in classification tasks exemplified by CPM and DeepIMV~\cite{CPM, DeepIMV}. \textbf{Imputation-Based Strategies.} Conversely, other methods focus on filling in the missing views using the available ones, followed by utilizing this complete dataset for downstream tasks. Simple imputation techniques include zero imputation, where missing values are replaced with zeros; mean imputation, which substitutes missing values with the mean of the available values; and K-means imputation, which leverages clustering to estimate and fill in missing values. More advanced techniques involve applying variational auto-encoders (VAEs) to generate estimates for the missing views in partial multi-view data~\cite{MVAE, MVTCAE, MIWAE}, as well as using generative adversarial networks (GANs) to reconstruct missing views~\cite{CPM-GAN, GAN_clu1, GAN_clu2}. Other approaches employ kernel canonical correlation analysis (CCA)\cite{MCWIV}, information theory principles~\cite{COMPLETER} and spectral graph theory\cite{IMVC} to perform imputation.

\subsection{Uncertainty-informed Deep Learning}
Deep neural networks have achieved remarkable success across various tasks, yet they often struggle to capture the uncertainty in their predictions, particularly when dealing with low-quality data~\cite{10086538}. Uncertainty in deep learning is generally divided into two types: aleatoric uncertainty, which pertains to the inherent noise in the data, and epistemic uncertainty, which relates to the uncertainty in the model itself. Methods for estimating uncertainty in deep learning~\cite{Gawlikowski2023survey} can be broadly categorized into four main types: single deterministic methods, Bayesian neural networks, ensemble methods, and test-time augmentation methods.

Single deterministic methods, such as the Evidential Deep Learning (EDL) approach~\cite{sensoy2018evidential}, compute category-specific evidence using a single deep neural network. Bayesian neural networks~\cite{gal2016dropout}, incorporate uncertainty estimates into the network parameters by treating them as probability distributions. Ensemble methods~\cite{lakshminarayanan2017simple} generate multiple models and aggregate their predictions to estimate uncertainty. Test-time augmentation methods~\cite{lyzhov2020greedy} use data augmentation techniques during inference to gauge uncertainty. Among these, EDL stands out for its unique approach to calculating evidence within a single deterministic framework, providing a direct way to quantify uncertainty in deep neural networks. This approach has proven effective in increasing the reliability and interpretability of model predictions.

Building on the strengths of EDL, the Evidential Deep Learning framework enhances model reliability by explicitly quantifying uncertainty, making it particularly useful in scenarios where understanding model confidence is crucial. Grounded in subjective logic principles \cite{josang2016subjective}, EDL addresses the challenges of model confidence and uncertainty highlighted in neural network calibration studies such as \cite{guo2017calibration}. The practical utility of EDL was further demonstrated by \cite{sensoy2018evidential}, who introduced methods to quantify classification uncertainty, thereby significantly increasing the trustworthiness of deep learning models. EDL's versatility has been showcased in applications like open set action recognition \cite{bao2021evidential}, illustrating its effectiveness in dealing with new and unseen data types. Its scope was further extended to multi-view classification \cite{han2021trusted}, demonstrating its ability to integrate and reason with information from multiple sources. This integration was enhanced by dynamic evidential fusion techniques \cite{han2022trusted}, underscoring EDL's adaptability in complex data environments. Recent advancements, such as adaptive EDL for semi-supervised learning \cite{yu2023adaptive} and its application in multimodal decision-making \cite{shao2024dual}, underscore EDL's progress in addressing real-world data challenges. Additionally, \cite{xu2024reliable} highlights EDL's potential in handling conflicting multi-view learning scenarios, reinforcing its capability to support reliable decision-making across diverse applications.

In the context of incomplete multi-view classification, existing methods rarely consider explicit measures of uncertainty. A few approaches, such as UIMC~\cite{xie2023exploring}, utilize EDL to measure the uncertainty of each view but overlook the potential conflicts when using the Dempster-Shafer theory during the fusion process. This can lead to counterintuitive and degraded performance due to the merging of conflicting evidence. To address this, we propose an Evidential Partial Multi-view Classification method that is robust to conflicts between views.

\section{Method}
\label{sec:method}
The proposed method aims to enhance robustness in decision-making using incomplete multi-view data by addressing uncertainty and potential conflicts. We define the problem of incomplete multi-view classification and discuss the limitations of existing methods (Section~\ref{sec:pro_definition}), discuss imputation methods for missing views (Section~\ref{sec:imputation}), and introduce a conflict-aware evidential aggregation approach (Section~\ref{sec:CAEA}) to integrate reliable decisions amidst imputation uncertainty. Finally, we briefly discuss the theoretical benefits of our approach (Section~\ref{sec:discuss}).

\begin{table}[ht]
\centering
\caption{A toy dataset: Basic Probability Assignments of Evidence.}
\begin{tabular}{cccccc}
\toprule
Evidence & \( m(A) \) & \( m(B) \) & \( m(C) \) & \( m(\{A, B, C\}) \) \\
\midrule
\( m_1 \) & 0.7        & 0.1        & 0.1        & 0.1                 \\
\( m_2 \) & 0.6        & 0.2        & 0.1        & 0.1                 \\
\( m_3 \) & 0.8        & 0.1        & 0.05       & 0.05                \\
\( m_4 \) & 0.1        & 0.8        & 0.05       & 0.05                \\
\( m_5 \) & 0.1        & 0.1        & 0.7        & 0.1                 \\
\bottomrule
\end{tabular}
\label{tab:bbas}
\end{table}

\begin{table}[ht]
\centering
\caption{Step-by-Step Fusion Results Using Dempster-Shafer Theory.}
\begin{tabular}{ccccc}
\toprule
Step              & \( m(A) \) & \( m(B) \) & \( m(C) \) & \( m(\{A, B, C\}) \) \\
\midrule
\( m_{1,2} \)       & 0.705      & 0.051      & 0.013      & 0.026               \\
\( m_{1,2,3} \)     & 0.745      & 0.040      & 0.010      & 0.035               \\
\( m_{1,2,3,4} \)   & 0.322      & 0.540      & 0.020      & 0.118               \\
\( m_{1,2,3,4,5} \) & 0.246      & 0.231      & 0.390      & 0.133               \\
\bottomrule
\end{tabular}
\label{tab:fusion-results}
\end{table}


\subsection{Preliminary}
\label{sec:pro_definition}
\textbf{Promblem definition.} In the context of multi-view classification, the objective is to establish a relationship between the training data and their labels by leveraging the diverse perspectives provided by multiple views. Consider $N$ training examples, each example $n$ represented by $\{\sX_n\}_{n=1}^N$ and comprising $V$ distinct views, expressed as $\sX = \{\mathbf{x}^v\}_{v=1}^{V}$. Each training example is associated with a class label from the set $\{\mathbf{y}_{n}\}_{n=1}^N$. This paper specifically addresses the issue of classification when some of the views are missing, as detailed in Def.~\ref{def:partial}.

\begin{figure*}
     \centering
      \includegraphics[width=\textwidth]{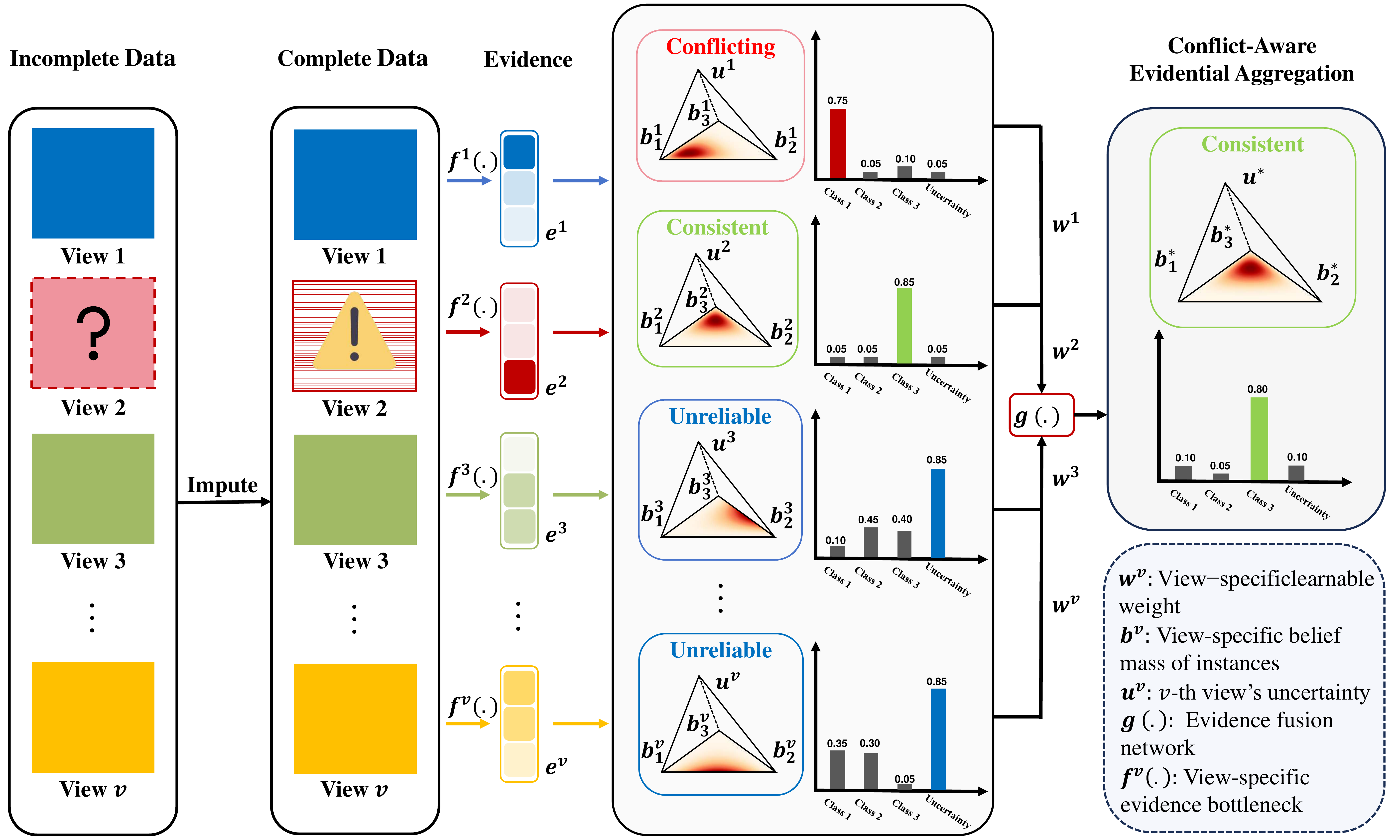}
\caption{Illustration of the proposed EDP-MVC. Initially, data with some missing views are identified and imputed to form complete multi-view data. Each view is then processed to extract evidence, including belief masses and uncertainties. The extracted evidence is evaluated for reliability, identifying conflicting, consistent, and unreliable views. To avoid unreliable fusion results and subsequent degradation, learnable weights, representing the reliability of each view, are used as discount factors for dynamic and adaptive evidence fusion. Through this dynamic reliability-driven discount fusion, the accuracy of uncertainty quantification is enhanced, and the reliability of decision-making is improved.}
\label{model}
\end{figure*}

\begin{definition}(Incomplete multi-view classification) \label{def:partial}
In a formal setting, a complete multi-view sample consists of $V$ views, denoted as $\sX = \{\mathbf{x}^v\}_{v=1}^{V}$, along with its corresponding class label $\mathbf{y}$. An incomplete multi-view observation, represented as $\overline{\sX}$, is a subset of the complete multi-view sample (i.e., $\overline{\sX} \subseteq \sX$) and can include any number of views, ranging from 1 to $V$ (i.e., $1 \leq \overline{V} \leq V$). Given a training dataset of $N$ samples, each consisting of an incomplete multi-view observation and its associated class label $\{\overline{\sX}_n, \mathbf{y}_n\}_{n=1}^{N}$, the task of incomplete multi-view classification involves learning a function that maps the incomplete multi-view observations $\overline{\sX}$ to their corresponding class labels $\mathbf{y}$.
\end{definition}

\textbf{Limitations of existing approaches.} In the realm of incomplete multi-view classification, two predominant strategies emerge: neglecting missing views entirely and imputing missing views. Methods that disregard missing views, such as those presented in \cite{DeepIMV, CPM}, solely rely on available data, often failing to exploit the inherent correlations between different views, which could potentially lead to a loss of valuable information. Conversely, imputation approaches attempt to reconstruct missing views using available data, as detailed in \cite{MVAE, MIWAE, impute_ADMM, impute_adversarial}. However, these methods frequently overlook the adverse effects that unreliable imputations can have on downstream tasks. A recent study \cite{xie2023exploring} proposes an innovative strategy to construct more reliable models by recognizing and leveraging the uncertainty associated with missing views, utilizing the Dempster-Shafer rule to integrate multi-view evidence. Despite its advancements, this approach fails to address potential conflicts among views, which can disrupt the fusion process and degrade overall performance. This issue is further compounded in the Dempster-Shafer combination framework, as evidenced in Tables~\ref{tab:bbas} and~\ref{tab:fusion-results}. These tables illustrate significant shifts in belief allocation following evidence fusion, underscoring the framework's sensitivity to conflicting evidence and often leading to an unreliable consensus that does not accurately represent the majority of the evidence.


\subsection{Imputing missing views}
\label{sec:imputation}
\textbf{Class Center Calculation.} For each view \( v \) and class \( c \), the class center \(\mathbf{c}_{v,c}\) is calculated as the average of the feature vectors \(\mathbf{x}_{v,i}\) for all samples \(i\) in class \(c\) with available data in view \(v\):
\begin{equation}
\mathbf{c}_{c}^{v} = \frac{1}{|I_{c}^{v}|} \sum_{i \in I_{v,c}} \mathbf{x}_{i}^{v}
\end{equation}
where \(I_{c}^{v}\) is the set of indices for samples in class \(c\) with available data in view \(v\), and \(|I_{c}^{v}|\) is the number of such samples.

\textbf{Imputation for Training Data.} For the training data, the imputed feature vector \(\tilde{\mathbf{x}}_{i}^{v}\) for a missing view \(v\) of \(i\)-th sample is given by the class center of the sample's class \(y_i\):
\begin{equation}
\label{eq:impute_train}
\centering
\tilde{\mathbf{x}}_{i}^{v} = \mathbf{c}_{y_i}^{v}
\end{equation}

\textbf{Imputation for Testing Data.} For the testing data, if view \(v\) is missing for a sample \(i\), the imputed feature vector \(\tilde{\mathbf{x}}_{i}^{v}\) is given by the class center that minimizes the sum of squared distances to the available views:
\begin{equation}
\label{eq:impute_test}
\tilde{\mathbf{x}}_{i}^{v} = \mathbf{c}_{c^*}^{v}, \quad \text{where } c^* = \arg\min_c \sum_{u \in U_i} \| \mathbf{x}_{i}^{u} - \mathbf{c}_{c}^{u} \|^2
\end{equation}
Here, \(U_i\) is the set of indices for views available for \(i\)-th sample .

\subsection{Conflict-aware multi-view learning}
\label{sec:CAEA}
The architecture depicted in Figure 3 integrates two main stages: view-specific evidential learning and evidential multi-view fusion. Initially, each view's incomplete data is imputed and processed through dedicated deep neural networks (DNNs), termed evidence bottlenecks \(f(\cdot)\), which generate evidence modeled as Dirichlet distributions. These distributions construct opinions, incorporating belief mass vectors and decision reliability, assessing conflicts in terms of potential view-specific uncertainty and overall consensus. By minimizing these conflicts, the DNNs are optimized to capture common multi-view information, thereby reducing decision discrepancies caused by view-specific errors. In the second stage, instead of a simple average pooling layer, the approach utilizes dynamic adaptive evidence fusion with learnable evidence discount factors. This method adjusts the integration of view-specific evidence dynamically, enhancing the system's consistency and reliability in decision-making across diverse and potentially conflicting views.

\subsubsection{View-specific evidential deep learning.}
In the initial stage of the proposed method, each view-specific DNN processes incomplete data, employing EDL \cite{sensoy2018evidential} to overcome the limitations of softmax-based DNNs in estimating predictive uncertainty. This approach integrates the evidence framework of Subjective Logic (SL) \cite{josang2016subjective}, where evidence from the input data supports the classification decisions. Specifically, we gather evidence, denoted as $\{\boldsymbol{\mathit{e}}_{n}^{v}\}$, using the view-specific DNNs $\{f^v(\cdot)\}_{v=1}^V$.

For a classification task with $K$ possible outcomes, a multinomial opinion for a specific view of an instance ($\mathbf{x}_{n}^{v}$) is represented by an ordered triplet $\mathbf{\mathit{w}} = (\boldsymbol{\mathit{b}}, \mathit{u}, \boldsymbol{\mathit{a}})$. Here, $\boldsymbol{\mathit{b}} = (\mathit{b}_{1}, \ldots, \mathit{b}_{k})^{\top}$ assigns belief masses to potential class outcomes based on the evidence, $\mathit{u}$ quantifies the uncertainty or vacuity in the evidence, and $\boldsymbol{\mathit{a}} = (\mathit{a}_{1}, \ldots, \mathit{a}_{k})^{\top}$ articulates the prior probability distribution across classes. The summation of belief and uncertainty masses is constrained by subjective logic to equal one:
\begin{align}
\sum_{k=1}^{K} \mathit{b}_{k} + \mathit{u} = 1, \quad \forall k \in [1, K],
\end{align}
ensuring non-negativity for $\mathit{b}_{k}$ and $\mathit{u}$.

The Dirichlet Probability Density Function (PDF), which is employed to model the category distributions, effectively captures second-order uncertainty, contrasting the first-order uncertainty represented by softmax probabilities. The Dirichlet PDF is expressed as:
\begin{align}
\mathit{D}(\mathbf{p}|\boldsymbol{\alpha}) = \left\{
\begin{matrix}
\frac{1}{\mathit{B}(\boldsymbol{\alpha})} \prod_{k=1}^{K} \mathit{p}_{k}^{\mathit{\alpha}_{k}-1}, & \text{for } \mathbf{p} \in \mathcal{S}_{K}, \\
0, & \text{otherwise},
\end{matrix}
\right.
\end{align}
where $\boldsymbol{\mathit{p}} = (\mathit{p}_{1}, \ldots, \mathit{p}_{k})^{\top}$ is the probability of class assignment, $\boldsymbol{\mathit{\alpha}} = (\mathit{\alpha}_{1}, \ldots, \mathit{\alpha}_{k})^{\top}$ denotes Dirichlet parameters, and $\mathcal{S}_{K}$ represents the $K$-dimensional unit simplex.

The parameters of the Dirichlet distribution, $\boldsymbol{\alpha}$, are determined by $\boldsymbol{\alpha} = \boldsymbol{\mathit{e}} + \textbf{1}$, ensuring all parameters exceed one for a non-sparse Dirichlet distribution. The relationship between the multinomial opinion and Dirichlet distribution is established as:
\begin{align}
\mathit{b}_{k} = \frac{\mathit{e}_{k}}{S} = \frac{\mathit{\alpha}_{k} - 1}{S}, \quad \mathit{u} = \frac{K}{S},
\end{align}
where $S = \sum_{k=1}^{K} (\mathit{e}_{k} + 1) = \sum_{k=1}^{K} \mathit{\alpha}_{k}$ is the total strength of the Dirichlet distribution. The level of uncertainty inversely correlates with the quantity of evidence, where minimal evidence equates to maximum uncertainty.

Finally, each view-specific DNN elucidates an opinion on the instance's category, which is then aggregated across views to form a comprehensive classification decision.

\subsubsection{Conflict-aware evidential aggregation}
This subsection discusses the challenges in fusing multi-view data, especially focusing on issues introduced by imputing missing data. In conflictive multi-view scenarios, noise views typically display high uncertainty, reducing their impact during fusion. Conversely, imputed views, attempting to compensate for missing data, might introduce biases that create unaligned views. These unaligned views provide highly conflicting opinions with low uncertainty, complicating the identification of reliable views and potentially leading to misalignments due to imputation rather than actual data discrepancies.

Indeed, the uncertainty in multi-view learning should correlate more with the quality of the views rather than merely increasing with their number. This is crucial when dealing with conflicts between imputed views, highlighting the need to assess each view's quality and alignment before fusion. Current methods often overlook these potential conflicts, which can stem from both noise and the biases inherent in the imputation process. Addressing these issues is vital for enhancing the reliability of multi-view data fusion, ensuring that strategies are robust against the misleading influences of poorly aligned views.

\begin{definition}
\label{def:op_aggregation}
Let \( \boldsymbol{w}^A = (b^A, u^A, a^A) \) and \( \boldsymbol{w}^B = (b^B, u^B, a^B) \) represent the opinions of views A and B over the same instance, respectively. The discount factors \( \gamma^A \) and \( \gamma^B \) adjust the respective evidences. The conflictive aggregated opinion \( \boldsymbol{w}^{A \underline{\Diamond} B} \) is recalculated as follows:
\begin{align}
\boldsymbol{w}^{A \underline{\Diamond} B} &= \boldsymbol{w}^A \underline{\Diamond} \boldsymbol{w}^B = (b^{A \underline{\Diamond} B}, u^{A \underline{\Diamond} B}, a^{A \underline{\Diamond} B}), \\
b_k^{A \underline{\Diamond} B} &= \frac{\gamma^A b_k^A u^B + \gamma^B b_k^B u^A}{u^A + u^B}, \\
u^{A \underline{\Diamond} B} &= \frac{2 u^A u^B}{u^A + u^B}, \\
a_k^{A \underline{\Diamond} B} &= \frac{\gamma^A a_k^A + \gamma^B a_k^B}{2}.
\end{align}
\end{definition}
To effectively manage the complexities and inherent biases of multi-view data fusion, particularly when views display variable levels of uncertainty and alignment issues, we introduce a refined approach to opinion aggregation. This method is crucial for synthesizing a coherent perspective from diverse and sometimes conflicting sources. Definition~\ref{def:op_aggregation} formalizes the conflict-aware evidential aggregation method, now enhanced with learnable discount factors. These factors dynamically adjust each view's contribution based on its reliability, allowing for a tailored integration of evidences. This method ensures each view's input is weighted by its evidential quality, thereby modulating the overall uncertainty when integrating views with high uncertainty. By implementing these adaptable discount factors, the aggregation process actively mitigates potential conflicts, avoiding overconfidence in the fused outcome and enhancing the decision-making process in environments with multi-view data.

\begin{definition}
\label{def:fusion}
Following Definition~\ref{def:op_aggregation}, we can fuse the final joint opinions $\boldsymbol{\mathit{w}}$ from different views as follows:
\begin{align}
\boldsymbol{\mathit{w}} = \boldsymbol{\mathit{w}}^{\mathit{1}} \underline{\Diamond} \boldsymbol{\mathit{w}}^{\mathit{2}} \underline{\Diamond} \ldots \underline{\Diamond} \boldsymbol{\mathit{w}}^{\mathit{V}}.
\end{align}
\end{definition}
Following this foundational method, Definition~\ref{def:fusion} extends the concept to multiple views. It outlines how to systematically apply the conflictive aggregation to fuse the final joint opinions from an arbitrary number of views. This sequential fusion process not only maintains consistency across different views but also scales effectively with the complexity and size of multi-view datasets.


\subsubsection{Loss function}

In this subsection, we discuss how traditional DNNs can be seamlessly converted into Evidential DNNs for obtaining multi-view joint opinions, an approach pioneered by Sensoy et al. \cite{sensoy2018evidential}. This conversion involves replacing the softmax output layer with a ReLU activation layer, treating the non-negative outputs as evidence, which informs the parameters of the Dirichlet distribution.

For an instance $\{\mathbf{x}_{n}^v\}_{v=1}^V$, the evidence vector $\mathbf{e}^{v}_{n} = f^{v}(\mathbf{x}^{v}_{n})$ predicted by the network is used to define the Dirichlet distribution parameters $\boldsymbol{\alpha}^{v}_{n} = \mathbf{e}^{v}_{n} + \mathbf{1}$.

Traditionally, neural networks utilize cross-entropy loss; however, in our proposed approach, this loss function is enhanced to integrate the parameters of the Dirichlet distribution, making it particularly well-suited for uncertainty-aware classification:
\begin{align}
\nonumber \mathcal{L}_{ece}(\boldsymbol{\alpha}_{n}) &= \int \left [ \sum_{j=1}^{K}-y_{nj}\log_{}{p_{nj}}\right ] \frac{\textstyle \prod_{j=1}^{K}p_{nj}^{{\alpha}_{nj}-1}}{B\left ( \boldsymbol{\alpha}_{n} \right)}d\mathbf{p}_{n}\\ 
&=\sum_{j=1}^{K}y_{nj}(\psi(S_{n})-\psi({\alpha}_{nj})),
\end{align}
where $\psi(\cdot)$ is the digamma function and $S_n = \sum_{j=1}^K \alpha_{nj}$ sums the evidence for instance $n$.

To ensure that incorrect labels generate lower evidence and discourage the model from assigning high confidence to incorrect classifications, we introduce an additional term in the loss function based on the Kullback-Leibler (KL) divergence as follows. It aims to penalize deviations from a target Dirichlet distribution that reflects true label probabilities, guiding the model to adjust its certainty levels based on actual performance. 
\begin{align}
\mathcal{L}_{KL}(\boldsymbol{\alpha}_{n}) &= \log \left(\frac{\Gamma(\sum_{k=1}^{K} \tilde{\alpha}_{nk})}{\Gamma(K) \prod_{k=1}^{K} \Gamma(\tilde{\alpha}_{nk})}\right) \\
&+ \sum_{k=1}^{K} (\tilde{\alpha}_{nk} - 1) \left[\psi(\tilde{\alpha}_{nk}) - \psi\left(\sum_{j=1}^{K} \tilde{\alpha}_{nj}\right)\right],
\end{align}
where $\tilde{\boldsymbol{\alpha}}_{n} = \mathbf{y}_{n} + (\mathbf{1} - \mathbf{y}_{n}) \odot \boldsymbol{\alpha}_{n}$ adjusts the parameters by removing non-misleading evidence.

The combined loss function is defined as:
\begin{align}
\label{eq:kl_loss}
\mathcal{L}_{acc}(\boldsymbol{\alpha}_{n}) = L_{ece}(\boldsymbol{\alpha}_{n}) + \lambda_{t} L_{KL}(\boldsymbol{\alpha}_{n}),
\end{align}
where $\lambda_{t} = \min(1.0, t/T)$ is an annealing coefficient that scales up the influence of KL divergence in the loss function over training epochs, indexed by $t$ up to a maximum of $T$. This controlled escalation prevents the premature stabilization of misclassified instances towards a uniform distribution, thereby ensuring that the model continues to refine its classifications and avoids settling on incorrect predictions too early in the training process.
To conclude, the comprehensive loss function for a specific instance $\{\mathbf{x}^v\}_{v=1}^{V}$ integrates several components and is given by:
\begin{align}
\label{eq:all_loss}
\mathcal{L} =  \mathcal{L}_{acc}(\overline{\boldsymbol{\alpha}_{n}}) + \beta \sum_{v=1}^{V} \mathcal{L}_{acc}(\boldsymbol{\alpha}_{n}^{v}).
\end{align}
where $\overline{\boldsymbol{\alpha}_{n}}$ represents the weighted fusion of evidences from different views using discounting factors.

\subsection{Discussion and analyses}
\label{sec:discuss}

In this subsection, we theoretically analyze the advantages of EDP-MVC, especially the conflictive opinion aggregation for the conflictive multi-view data. The following propositions provide the theoretical analysis to support the conclusions.

\textbf{Proposition 1.} The conflictive opinion aggregation $\boldsymbol{\mathit{w} } ^{A\underline{\Diamond } B}$ is equivalent to aggregate the view-specific evidences. It updates the conflictive opinion aggregation formula to include learnable discount factors, resulting in a more sophisticated method for integrating view-specific evidences as below.
\begin{equation}
  \boldsymbol{\mathit{e}}^{A\underline{\Diamond} B} = \gamma^A \boldsymbol{\mathit{e}}^{\mathit{A}} + \gamma^B \boldsymbol{\mathit{e}}^{\mathit{B}},  
\end{equation}
where \( \gamma^A \) and \( \gamma^B \) denote the discount factors for views A and B, optimized during training to reflect the reliability of each view.
Therefore, in the multi-view fusion stage, we establish a simple yet effective weighted average pooling fusion layer, 
$g(\cdot)$, to achieve conflict-aware evidential aggregation. It exhibit some potential benefits as follows:
\begin{itemize}
    \item \textbf{Enhanced conflict resolution.} Prioritizes more credible sources, reducing the impact of unreliable or noisy views, thereby improving the integrity of the fusion process.
    \item \textbf{Dynamic adaptability.} Allows the model to adjust to changes in data quality or view reliability, enhancing robustness against varying data conditions.
    \item \textbf{Improved accuracy.} Focuses on the quality of data rather than quantity, ensuring more accurate and reliable decision-making.
    \item \textbf{Customizability and scalability.} The learnable factors facilitate fine-tuning for specific applications and scalability to accommodate additional views.
\end{itemize}

\textbf{Proposition 2} articulates that in conflictive opinion aggregation, the uncertainty mass of the aggregated opinion is directly influenced by the uncertainty of the newly integrated opinion. Specifically, if the uncertain mass of the new opinion is smaller than that of the original, the uncertainty of the aggregated opinion decreases; conversely, if it is larger, the uncertainty increases.

Existing incomplete multi-view classification methods, (\emph{e.g.} \cite{xie2023exploring}), often assume that the fusion of opinions will invariably lead to a reduction in uncertainty. This assumption fails to consider the nature of the evidence being integrated: reliable, corroborating evidence should indeed diminish uncertainty, while conflicting or unreliable evidence should justifiably increase it. The discrepancies often arise from the inherent uncertainty in imputed data and misalignment, as well as significant variances across different views.

Consider an illustrative example involving two sensors monitoring environmental conditions in an area prone to frequent atmospheric changes. Sensor A is highly sensitive and can detect subtle changes in conditions, while Sensor B, less sensitive, occasionally misreads data, especially under fluctuating conditions. When both sensors report stable conditions, their corroborated evidence can be fused to reduce uncertainty effectively. However, if Sensor A detects a sudden atmospheric change which Sensor B fails to register, the fusion of this conflicting data should increase the uncertainty, highlighting the disparity in sensor reliability.

In such cases, rather than indiscriminately reducing uncertainty, fusion methods need to incorporate mechanisms that adjust the weight of each view's contribution based on its reliability. This approach ensures that the fusion process does not merely average out the information but evaluates and scales it based on the quality of evidence each view provides. 

To summarize, existing fusion methods in multi-view classification should account for potential conflicts and the variable reliability of the data from different views. Each piece of evidence should be critically assessed and integrated in a manner that reflects its informational value, rather than assuming a uniform reduction in uncertainty. The proposed EDP-MVC aims to solve this issue and make a difference in scenarios where data completeness and quality are variable. The pseudo-code of EDP-MVC is provided in Alg.~\ref{alg:code_train}.

\begin{algorithm}[!htbp]
\footnotesize
\caption{Training Process of EDP-MVC}
\label{alg:code_train}
\SetAlgoLined
\textbf{Input:} Incomplete multi-view training set $\{\overline{\displaystyle \sX}_n, \mathbf{y}_n\}_{n=1}^N$.\\
\textbf{Initialize:}
Initialize the parameters of EDP-MVC and learnable view-specific evidence discount factors $\{\gamma^v\}_{v=1}^{V}$.\\
\For{$n=1$ \textbf{to} $N$}{
    \For{$m=1$ \textbf{to} $M$ \textbf{where} $x_n^m$ is missing}{
        Impute missing views in both training and testing data using Eqs.~(\ref{eq:impute_train}) and (\ref{eq:impute_test}).\\
    }
}
\While{not converged}{
    \For{$n=1$ \textbf{to} $N$}{
        \For{$v=1$ \textbf{to} $V$}{
            Extract evidence ${\boldsymbol{e}}_n^v$ using the DNN for view $v$;\\
            Compute $\boldsymbol{\alpha}_n^v = {\boldsymbol{e}}_n^v + 1$;\\
            Derive $\mathcal{S}_n^v=\{\{b_{n,k}^v\}_{k=1}^K, u_n^v, a_n^v\}$ based on the evidence;\\
        }
        Aggregate the evidences across views to compute ${\boldsymbol{\alpha}}_{n} = [\alpha_{n,1}, \cdots, \alpha_{n,K}]$ and $\mathcal{S}_n = \{\{b_{n,k}\}_{k=1}^K, u_n, a_n^v\}$ using Definition~\ref{def:op_aggregation};\\
    }
    Calculate the overall loss $\mathcal{L}$ using Eq.~(\ref{eq:all_loss});\\
    Update the model parameters and $\{\gamma^v\}_{v=1}^{V}$ using gradient descent or an appropriate optimization algorithm;\\
}
\textbf{Output:} Trained model parameters.\\
\end{algorithm}

\section{Experiments}
\label{sec:Experiments}
In this section, we conduct comprehensive experiments across multiple multi-view datasets characterized by various missing rates $\eta = \frac{\sum_{v=1}^{V}{M}^v}{V \times N}$, where $M^v$ denotes the number of observations lacking the $v$-th view.

The research is meticulously structured to address the ensuing pivotal inquiries:

\textbf{Q1 Effectiveness (I).} Does the proposed method exhibit superior performance compared to alternative approaches?

\textbf{Q2 Effectiveness (II).} How does the proposed method's performance align with that of the latest state-of-the-art methods when applied to larger datasets?

\textbf{Q3 Ablation study.} Within the context of varying missing view rates and conflicting data, how do strategies based on Dempster-Shafer theory for multi-view evidence integration fare?

\textbf{Q4 Parameter sensitivity.} How do the hyperparameters within the Eq.~\ref{eq:kl_loss} impact the experimental outcomes?
\begin{table}[htbp]
\centering
\caption{Overview of real-world datasets used in experiments}
\label{tab:datasets}
\scriptsize 
\begin{tabular}{@{}lccc@{}}
\toprule
\textbf{Dataset} & \textbf{Size} & \textbf{Classes} & \textbf{Dimensionality} \\ \midrule
YaleB~\cite{YaleB} & 650 & 10 & 2500\textbar3304\textbar6750 \\
ROSMAP~\cite{ROSMAP} & 351 & 2 & 200\textbar200\textbar200 \\
Handwritten~\cite{Handwritten} & 2000 & 10 & 240\textbar76\textbar216\textbar47\textbar64\textbar6 \\
BRCA~\cite{ROSMAP} & 875 & 5 & 1000\textbar1000\textbar503 \\
Scene15~\cite{Scene} & 4485 & 15 & 20\textbar59\textbar40 \\
Caltech-101~\cite{Caltech} & 9144 & 102 & 48\textbar40\textbar254\textbar1984\textbar512\textbar928 \\
NUS-WIDE-OBJECT~\cite{NUS} & 30000 & 31 & 129\textbar74\textbar145\textbar226\textbar65 \\ \bottomrule
\end{tabular}
\end{table}

\begin{table*}[htbp]
\centering
\caption{Comparison in terms of classification accuracy (mean$\pm$std) with $\eta = [0, 0.1, 0.2, 0.3, 0.4, 0.5]$ on five datasets.} 
\label{tab:acc} 
\scalebox{0.94}{
\begin{tabular}{c|c|ccccccc}
\toprule
\multirow{2}{*}{Datasets}    & \multirow{2}{*}{Methods} & \multicolumn{6}{c}{Missing rates}                                                                         \\
                             &                          & $\eta= 0$               & $\eta=0.1$             & $\eta=0.2$             &$\eta=0.3$             & $\eta=0.4$             & $\eta=0.5$             \\
\midrule
\multirow{8}{*}{YaleB}       & Mean-Imputation                    & $\pmb{{1.0000\pm0.00}}$ & $0.9923\pm0.00$          & $0.9769\pm0.01$          & $0.9769\pm0.01$          & $0.9692\pm0.01$          & $0.9615\pm0.01$          \\
                             & GCCA                     & $0.9692\pm0.00$          & $0.9385\pm0.01$          & $0.9077\pm0.02$          & $0.8615\pm0.03$          & $0.8385\pm0.02$          & $0.8231\pm0.02$          \\
                             & TCCA                     & $0.9846\pm0.00$          & $0.9625\pm0.00$          & $0.9492\pm0.01$          & $0.9077\pm0.01$          & $0.8846\pm0.02$          & $0.8615\pm0.01$          \\
                             & MVAE                     & $\pmb{1.0000\pm0.00}$ & $0.9969\pm0.00$          & $0.9861\pm0.00$          & $0.9831\pm0.01$          & $0.9692\pm0.02$          & $0.9599\pm0.01$          \\
                             & MIWAE                    & $\pmb{1.0000\pm0.00}$ & $0.9923\pm0.00$          & $0.9923\pm0.01$          & $0.9903\pm0.01$          & $0.9846\pm0.01$          & $0.9692\pm0.03$          \\
                             & CPM-Nets                      & $0.9915\pm0.02$          & $0.9862\pm0.01$          & $0.9800\pm0.01$          & $0.9700\pm0.02$          & $0.9469\pm0.01$          & $0.9100\pm0.02$          \\
                             & DeepIMV                  & $\pmb{1.0000\pm0.00}$ & $0.9846\pm0.03$          & $0.9231\pm0.02$          & $0.9154\pm0.08$          & $0.8923\pm0.02$          & $0.8718\pm0.06$          \\ 
                          & DCP                  & $0.9938\pm0.00$          & $0.9803\pm0.01$          & $0.9682\pm0.03$          & $0.9350\pm0.01$          & $0.9134\pm0.02$          & $0.8902\pm0.02$          \\                             
                             & UIMC                      & $\pmb{1.0000\pm0.00}$ & $\pmb{1.0000\pm0.00}$ & $\pmb{0.9981\pm0.00}$ & $\pmb{0.9962\pm0.01}$ & $\pmb{0.9847\pm0.01}$ & $\pmb{0.9769\pm0.01}$ \\\cmidrule{2-8}
                             & Ours                      & $\pmb{1.0000\pm0.00}$ & $\pmb{1.0000\pm0.00}$ & $0.9923\pm0.00$ & $0.9692\pm0.01$ & $0.9538\pm0.00$ & $0.9462\pm0.00$\\  
\midrule
\multirow{8}{*}{ROSMAP}      & Mean-Imputation                    & $0.7429\pm0.03$          & $0.6809\pm0.01$          & $0.6714\pm0.02$          & $0.6571\pm0.07$          & $0.6429\pm0.02$          & $0.6072\pm0.05$          \\
                             & GCCA                     & $0.6953\pm0.03$          & $0.6571\pm0.02$          & $0.6429\pm0.04$          & $0.6143\pm0.03$          & $0.5714\pm0.06$          & $0.5429\pm0.06$          \\
                             & TCCA                     & $0.7143\pm0.03$          & $0.7072\pm0.01$          & $0.6857\pm0.05$          & $0.6500\pm0.05$          & $0.6286\pm0.03$          & $0.6036\pm0.06$          \\
                             & MVAE                     & $0.7429\pm0.02$          & $0.7286\pm0.05$          & $0.7143\pm0.05$          & $0.6786\pm0.03$          & $0.6786\pm0.06$          & $0.6524\pm0.06$          \\
                             & MIWAE                    & $0.7429\pm0.03$          & $0.7286\pm0.02$          & $0.6714\pm0.05$          & $0.6571\pm0.03$          & $0.6571\pm0.05$          & $0.6357\pm0.08$          \\
                             & CPM-Nets                      & $0.7840\pm0.05$          & $0.7517\pm0.04$          & $0.7394\pm0.06$          & $0.7183\pm0.04$          & $0.6901\pm0.08$          & $0.6409\pm0.08$          \\
                             & DeepIMV                  & $0.7607\pm0.03$          & $0.7429\pm0.01$          & $0.7143\pm0.05$          & $0.6643\pm0.05$          & $0.6524\pm0.06$          & $0.6250\pm0.06$          \\
                          & DCP                  & $0.7353\pm0.06$          & $0.7029\pm0.05$          & $0.6863\pm0.05$          & $0.6735\pm0.06$          & $0.6347\pm0.04$          & $0.6031\pm0.05$          \\                             
                             & UIMC                      & $0.8714\pm0.00$ & $0.8429\pm0.03$ & $0.7714\pm0.05$ & $0.7464\pm0.03$ & $0.7214\pm0.03$ & $0.7143\pm0.02$ \\\cmidrule{2-8}  
                             & Ours                      & $\pmb{0.8857\pm0.00}$ & $\pmb{0.8429\pm0.00}$ & $\pmb{0.8287\pm0.00}$ & $\pmb{0.7857\pm0.00}$ & $\pmb{0.7571\pm0.00}$ & $\pmb{0.7239\pm0.01}$\\                           
\midrule
\multirow{8}{*}{Handwritten} & Mean-Imputation                    & $0.9800\pm0.00$          & $0.9750\pm0.00$          & $0.9700\pm0.01$          & $0.9700\pm0.01$          & $0.9500\pm0.01$          & $0.9100\pm0.01$          \\
                             & GCCA                     & $0.9500\pm0.01$          & $0.9350\pm0.02$          & $0.9100\pm0.01$          & $0.8875\pm0.02$          & $0.8425\pm0.02$          & $0.8200\pm0.03$          \\
                             & TCCA                     & $0.9725\pm0.00$          & $0.9650\pm0.00$          & $0.9575\pm0.02$          & $0.9350\pm0.01$          & $0.9200\pm0.01$          & $0.9100\pm0.02$          \\
                             & MVAE                     & $0.9800\pm0.00$          & $0.9750\pm0.01$          & $0.9700\pm0.00$          & $0.9650\pm0.01$          & $0.9575\pm0.01$          & $0.9500\pm0.01$          \\
                             & MIWAE                    & $0.9800\pm0.00$          & $0.9800\pm0.00$         & $0.9725\pm0.00$          & $0.9650\pm0.00$          & $0.9475\pm0.01$          & $0.9375\pm0.02$         \\
                             & CPM-Nets                      & $0.9550\pm0.01$          & $0.9475\pm0.01$          & $0.9375\pm0.01$          & $0.9300\pm0.02$          & $0.9225\pm0.01$          & $0.9125\pm0.01$          \\
                             & DeepIMV                  & $\pmb{0.9908\pm0.04}$ & $\pmb{0.9883\pm0.02}$ & $\pmb{0.9850\pm0.04}$ & $0.9750\pm0.02$ & $0.9625\pm0.04$          & $0.9450\pm0.06$          \\ 
                          & DCP                  & $0.9715\pm0.01$          & $0.9585\pm0.01$          & $0.9425\pm0.01$          & $0.9350\pm0.01$          & $0.9125\pm0.02$          & $0.8950\pm0.02$          \\                             
                             & UIMC                      & $0.9825\pm0.00$          & $0.9800\pm0.00$         & $0.9800\pm0.00$          & $\pmb{0.9775\pm0.00}$ & $0.9700\pm0.01$ & $0.9600\pm0.01$ \\\cmidrule{2-8}
                             & Ours                      & $0.9850\pm0.00$ & $0.9850\pm0.01$ & $0.9775\pm0.00$ & $\pmb{0.9775\pm0.00}$ & $\pmb{0.9700\pm0.00}$ & $\pmb{0.9625\pm0.01}$\\                             
\midrule
\multirow{8}{*}{BRCA}        & Mean-Imputation                    & $0.7885\pm0.02$          & $0.7143\pm0.03$          & $0.7000\pm0.04$          & $0.6571\pm0.02$          & $0.6429\pm0.03$          & $0.6286\pm0.02$          \\
                             & GCCA                     & $0.7371\pm0.03$          & $0.7143\pm0.03$          & $0.6971\pm0.04$          & $0.6762\pm0.02$          & $0.6514\pm0.03$          & $0.6381\pm0.04$          \\
                             & TCCA                     & $0.7543\pm0.02$          & $0.7314\pm0.03$          & $0.7238\pm0.04$          & $0.7129\pm0.03$          & $0.6857\pm0.04$          & $0.6743\pm0.03$          \\
                             & MVAE                     & $0.7885\pm0.03$ &$0.7691\pm0.02$ & $0.7347\pm0.01$          & $0.6968\pm0.03$          & $0.6633\pm0.05$          & $0.6388\pm0.03$          \\
                             & MIWAE                    & $0.7885\pm0.02$          & $0.7352\pm0.03$          & $0.7314\pm0.03$          & $0.7105\pm0.02$          & $0.7029\pm0.02$          & $0.6857\pm0.04$          \\
                             & CPM-Nets                      & $0.7388\pm0.02$          & $0.7317\pm0.04$          & $0.7107\pm0.08$          & $0.7233\pm0.04$          & $0.6980\pm0.05$          & $0.6788\pm0.03$          \\
                             & DeepIMV                  & $0.7686\pm0.03$          & $0.7614\pm0.02$          & $0.7457\pm0.02$          & $0.7414\pm0.02$          & $0.7400\pm0.02$          & $0.6714\pm0.04$          \\
                          & DCP                  & $0.7773\pm0.01$          & $0.7741\pm0.02$          & $0.7598\pm0.02$          & $0.7452\pm0.01$          & $0.7323\pm0.01$          & $0.6973\pm0.02$          \\                             
                             & UIMC                      & $0.8286\pm0.01$ & $0.7943\pm0.01$          & $0.7771\pm0.01$ & $0.7657\pm0.02$ & $\pmb{0.7543\pm0.02}$ & $\pmb{0.7429\pm0.02}$ \\\cmidrule{2-8}
                             & Ours                      & $\pmb{0.8453\pm0.01}$ & $\pmb{0.8229\pm0.00}$ & $\pmb{0.8057\pm0.01}$ & $\pmb{0.8000\pm0.01}$ & $0.7427\pm0.01$ & $0.7086\pm0.01$\\
\midrule
\multirow{8}{*}{Scene15}       & Mean-Imputation                    & $0.7681\pm0.02$          & $0.6912\pm0.01$          & $0.6477\pm0.01$          & $0.6098\pm0.01$          & $0.5864\pm0.02$          & $0.5106\pm0.02$          \\
                             & GCCA                     & $0.6611\pm0.02$          & $0.6511\pm0.01$          & $0.6176\pm0.01$          & $0.5708\pm0.01$          & $0.5385\pm0.02$          & $0.5006\pm0.02$          \\
                             & TCCA                     & $0.6878\pm0.02$          & $0.6644\pm0.01$          & $0.6566\pm0.01$          & $0.6187\pm0.01$          & $0.5741\pm0.01$          & $0.5563\pm0.02$          \\
                             & MVAE                     & $0.7681\pm0.00$ & $0.7346\pm0.01$ & $0.7157\pm0.01$ & $0.6689\pm0.01$          & $0.6444\pm0.01$          & $0.6098\pm0.01$          \\
                             & MIWAE                    & $0.7681\pm0.03$          & $0.7179\pm0.01$          & $0.6990\pm0.01$         & $0.6566\pm0.01$          & $0.6265\pm0.02$          & $0.5875\pm0.02$          \\
                             & CPM-Nets                      & $0.6990\pm0.02$          & $0.6566\pm0.02$          & $0.6388\pm0.00$          & $0.6265\pm0.01$          & $0.5903\pm0.01$          & $0.5708\pm0.01$          \\
                             & DeepIMV                  & $0.7124\pm0.00$          & $0.6934\pm0.02$          & $0.6656\pm0.01$          & $0.6410\pm0.00$          & $0.5853\pm0.02$          & $0.5719\pm0.01$          \\
                          & DCP                  & $0.7510\pm0.01$          & $0.7337\pm0.01$          & $0.7122\pm0.01$          & $0.7124\pm0.01$          & $0.6907\pm0.01$          & $\pmb{0.6570\pm0.01}$          \\                            
                             & UIMC                      & $0.7770\pm0.00$ & $0.7581\pm0.01$ & $0.7347\pm0.00$ & $0.6990\pm0.01$ & $0.6689\pm0.01$ & $0.6254\pm0.02$\\\cmidrule{2-8}
                             & Ours                      & $\pmb{0.7984\pm0.00}$ & $\pmb{0.7871\pm0.01}$ & $\pmb{0.7759\pm0.00}$ & $\pmb{0.7492\pm0.01}$ & $\pmb{0.6958\pm0.01}$ & $0.6287\pm0.00$\\
                             
\bottomrule
\end{tabular}}
\end{table*}

\subsection{Datasets and comparison methods}
\label{sec:dataset}

\begin{table*}
\centering
\caption{Comparison in terms of classification accuracy with noisy views (mean$\pm$std) with $\eta = [0, 0.1, 0.2, 0.3, 0.4, 0.5]$ on five datasets.}
\label{tab:noise}
\resizebox{1.0\textwidth}{!}{
\begin{tabular}{c|c|ccccccc}
\toprule
\multirow{2}{*}{Datasets}    & \multirow{2}{*}{Methods} & \multicolumn{6}{c}{Missing rates}                                                                         \\
                             &                          & $\eta= 0$               & $\eta=0.1$             & $\eta=0.2$             &$\eta=0.3$             & $\eta=0.4$             & $\eta=0.5$             \\
\midrule
\multirow{3}{*}{YaleB}       & UIMC                      & $\pmb{1.0000\pm0.00}$ & $0.9923\pm0.00$ & $0.9846\pm0.00$ & $0.9308\pm0.01$ & $0.8462\pm0.01$ & $0.6769\pm0.02$ \\
                             & Naive Fusion (Ours)                      & $\pmb{1.0000\pm0.00}$ & $0.9923\pm0.00$ & $0.9615\pm0.00$ & $0.9308\pm0.01$ & $0.8923\pm0.00$ & $0.8538\pm0.00$\\\cmidrule{2-8}
                             & Discount Fusion (Ours)                      & $\pmb{1.0000\pm0.00}$ & $\pmb{1.0000\pm0.00}$ & $\pmb{0.9923\pm0.00}$ & $\pmb{0.9769\pm0.01}$ & $\pmb{0.9615\pm0.00}$ & $\pmb{0.9000\pm0.01}$\\  
\midrule
\multirow{3}{*}{ROSMAP}      & UIMC                      & $0.7529\pm0.01$ & $0.7429\pm0.00$ & $0.7286\pm0.02$ & $0.6983\pm0.02$ & $0.6714\pm0.01$ & $0.5571\pm0.00$ \\
                             & Naive Fusion (Ours)                      & $0.7714\pm0.01$ & $0.7571\pm0.00$ & $0.7381\pm0.01$ & $0.7229\pm0.01$ & $0.6571\pm0.01$ & $0.6429\pm0.01$\\\cmidrule{2-8}  
                             & Discount Fusion (Ours)                      & $\pmb{0.8286\pm0.01}$ & $\pmb{0.8142\pm0.01}$ & $\pmb{0.7571\pm0.00}$ & $\pmb{0.7429\pm0.00}$ & $\pmb{0.7286\pm0.00}$ & $\pmb{0.6857\pm0.00}$\\                           
\midrule
\multirow{3}{*}{Handwritten} & UIMC                      & $0.9825\pm0.00$          & $0.9700\pm0.00$         & $0.9500\pm0.00$          & $0.9400\pm0.00$ & $0.9000\pm0.01$ & $0.8350\pm0.02$ \\
                             & Naive Fusion (Ours)                     & $0.9825\pm0.00$ & $0.9775\pm0.00$ & $0.9550\pm0.00$ & $0.9350\pm0.00$ & $0.9125\pm0.01$ & $0.8725\pm0.01$\\\cmidrule{2-8}
                             & Discount Fusion (Ours)                      & $\pmb{0.9875\pm0.00}$ & $\pmb{0.9800\pm0.00}$ & $\pmb{0.9775\pm0.00}$ & $\pmb{0.9700\pm0.00}$ & $\pmb{0.9625\pm0.00}$ & $\pmb{0.9550\pm0.01}$\\                             
\midrule
\multirow{3}{*}{BRCA}        & UIMC                      & $0.7371\pm0.01$ & $0.6914\pm0.02$          & $0.6800\pm0.01$ & $0.6686\pm0.01$ & $0.6571\pm0.01$ & $0.5771\pm0.02$ \\
                             & Naive Fusion (Ours)                      & $0.7943\pm0.00$ & $0.7714\pm0.00$ & $0.7657\pm0.01$ & $0.7314\pm0.01$ & $0.6457\pm0.02$ & $0.5543\pm0.01$\\\cmidrule{2-8}
                             & Discount Fusion (Ours)                      & $\pmb{0.8057\pm0.00}$ & $\pmb{0.7829\pm0.00}$ & $\pmb{0.7714\pm0.01}$ & $\pmb{0.7429\pm0.01}$ & $\pmb{0.7257\pm0.01}$ & $\pmb{0.7098\pm0.01}$\\
\midrule
\multirow{3}{*}{Scene15}       & UIMC                      & $0.7157\pm0.01$ & $0.6979\pm0.01$ & $0.6711\pm0.01$ & $0.6210\pm0.01$ & $0.5530    \pm0.04$ & $0.4994\pm0.01$\\
                             & Naive Fusion (Ours)                      & $0.7648\pm0.00$ & $0.7302\pm0.01$ & $0.6656\pm0.00$ & $0.6132\pm0.01$ & $0.5474\pm0.01$ & $0.4604\pm0.00$\\\cmidrule{2-8}
                             & Discount Fusion (Ours)                      & $\pmb{0.7741\pm0.00}$ & $\pmb{0.7447\pm0.01}$ & $\pmb{0.6867\pm0.00}$ & $\pmb{0.6221\pm0.01}$ & $\pmb{0.5696\pm0.01}$ & $\pmb{0.5396\pm0.00}$\\
\bottomrule
\end{tabular}}
\end{table*}

\textbf{Real-world datasets.} To evaluate the effectiveness of our proposed method, we conducted experiments on five medium-scale datasets and two large-scale multi-view datasets as shown in Table~\ref{tab:datasets}. Specifically, \textbf{YaleB}~\cite{YaleB} is a 3-view dataset with 10 categories, each containing 65 facial images. \textbf{ROSMAP}~\cite{ROSMAP} is a 3-view dataset with two categories: Alzheimer's disease (AD) patients (182 samples) and normal controls (NC) (169 samples). \textbf{Handwritten}~\cite{Handwritten} is a 6-view dataset with 10 categories (digits 0 to 9), each containing 200 samples. \textbf{BRCA}~\cite{ROSMAP} is a 3-view dataset for Breast Invasive Carcinoma (BRCA) subtype classification, with 5 categories and 46 to 436 samples per category. \textbf{Scene15}~\cite{Scene} is a 3-view dataset for scene classification, with 15 categories and 210 to 410 samples per category. We also conducted experiments on two large-scale multi-view datasets. \textbf{Caltech-101}~\cite{Caltech} is a 6-view dataset consisting of images from 101 object classes, plus one background clutter class, totaling 9144 images, with each class containing approximately 40 to 800 images. \textbf{NUS-WIDE-OBJECT}~\cite{NUS} is a 5-view dataset with 31 classes and a total of 30,000 images.

\textbf{Conflictive datasets.} To further challenge the robustness of the proposed incomplete multi-view classification method under the Dempster-Shafer theory, we introduce scenarios with misaligned and noisy views based on the real-world datasets. Specifically, for misaligned views, a portion of the dataset is modified such that for a given sample, the view data of class $c$ is confused with that of the next class (\emph{i.e.} $(c+1) \mod C$, where $C$ denotes the total number of class). Noisy views are simulated by adding three types of noise views to the original dataset: views consisting entirely of zeros, ones, and random noise (\emph{i.e.} Gaussian Noise). Through these adversarial experiments, we aim to evaluate the robustness of our approach.

\textbf{Comparison methods.} We compared our proposed method with the following approaches:
(1) \textbf{Mean-Imputation}: Imputes the missing view $\mathbf{x}^m$ using the mean of all available observations in the $m$-th view.
(2) \textbf{GCCA}~\cite{GCCA}: Extends Canonical Correlation Analysis (CCA)~\cite{CCA} to handle data with more than two views.
(3) \textbf{TCCA}~\cite{TCCA}: Maximizes the canonical correlations among multiple views to find a common subspace shared by all views.
(4) \textbf{MVAE}~\cite{MVAE}: Extends the variational autoencoder to multi-view data, employing a product-of-experts strategy to obtain a shared latent subspace.
(5) \textbf{MIWAE}~\cite{MIWAE}: Extends the importance-weighted autoencoder to multi-view data to impute missing values.
(6) \textbf{CPM-Nets}~\cite{CPM}: Learns joint latent representations for all views with available data and maps these representations to classification predictions.
(7) \textbf{DeepIMV}~\cite{DeepIMV}: Utilizes the information bottleneck framework to derive marginal and joint representations from available data, constructing view-specific and multi-view predictors for classification.
(8) \textbf{DCP}~\cite{Lin2023DCP}: Provides an information-theoretic framework integrating consistency learning and data recovery, imputing missing views by minimizing conditional entropy through dual prediction.
(9) \textbf{UIMC}~\cite{xie2023exploring}: Samples a Gaussian distribution from the complete samples to impute the missing data. Specifically, it constructs the nearest neighbor set from this distribution to create imputed datasets. These imputed multi-view evidences are then fused using the Dempster-Shafer rule, ensuring robust and effective integration of quality opinions from the multi-view imputed data.

\subsection{Experiment results}
\begin{figure}[htbp]
     \centering
      \includegraphics[width=0.5\textwidth]{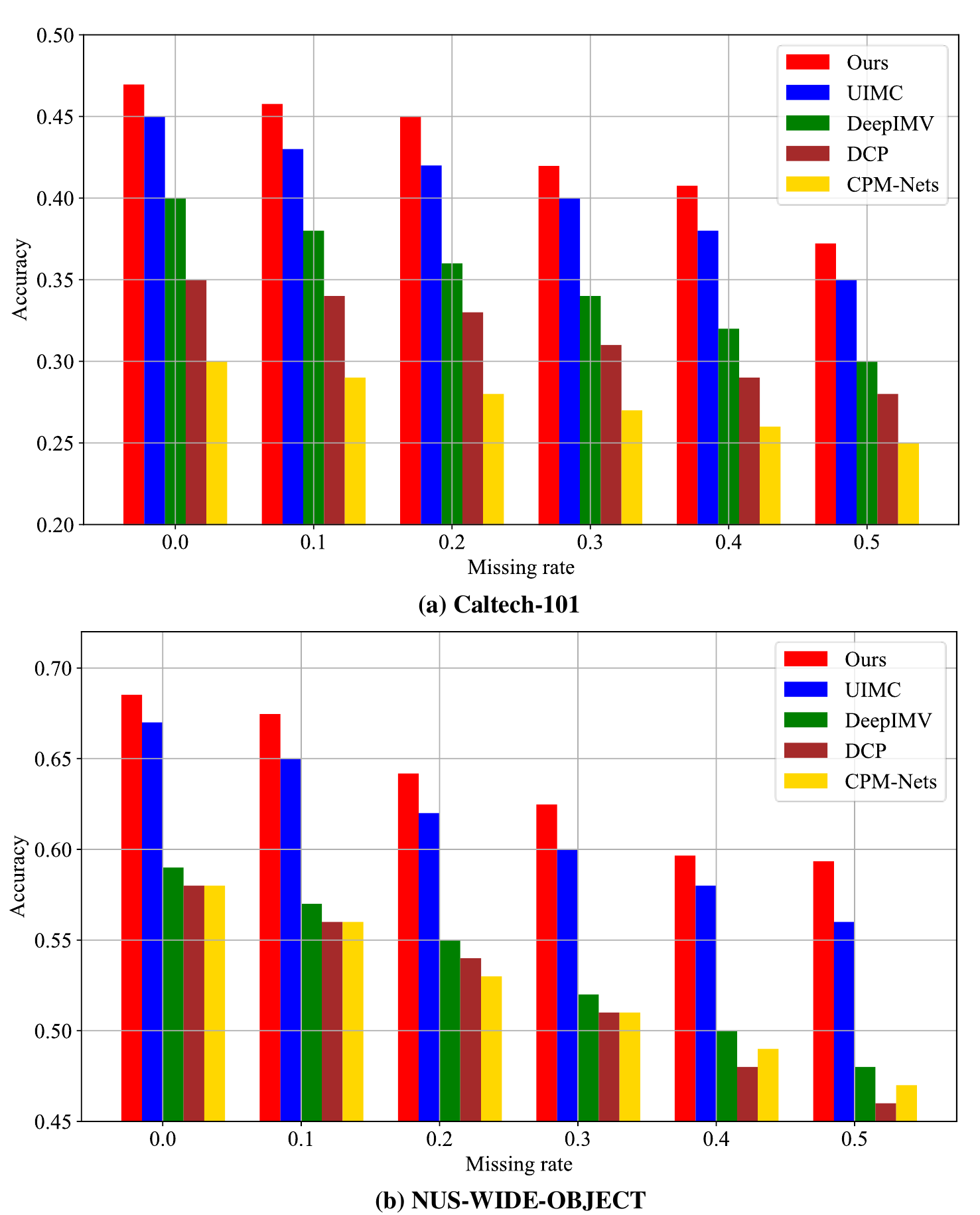}
\caption{Classification performance with varied missing rates. The results demonstrate that our proposed method (Ours) consistently outperforms other cutting-edge methods (UIMC, DeepIMV, DCP, and CPM-Nets) on large-scale datasets, including Caltech-101 and NUS-WIDE-OBJECT. This superior performance highlights the robustness and effectiveness of our approach in handling large and complex incomplete datasets. }
\label{fig:large_scale}
\end{figure}
\begin{figure}[htbp]
     \centering
      \includegraphics[width=0.5\textwidth]{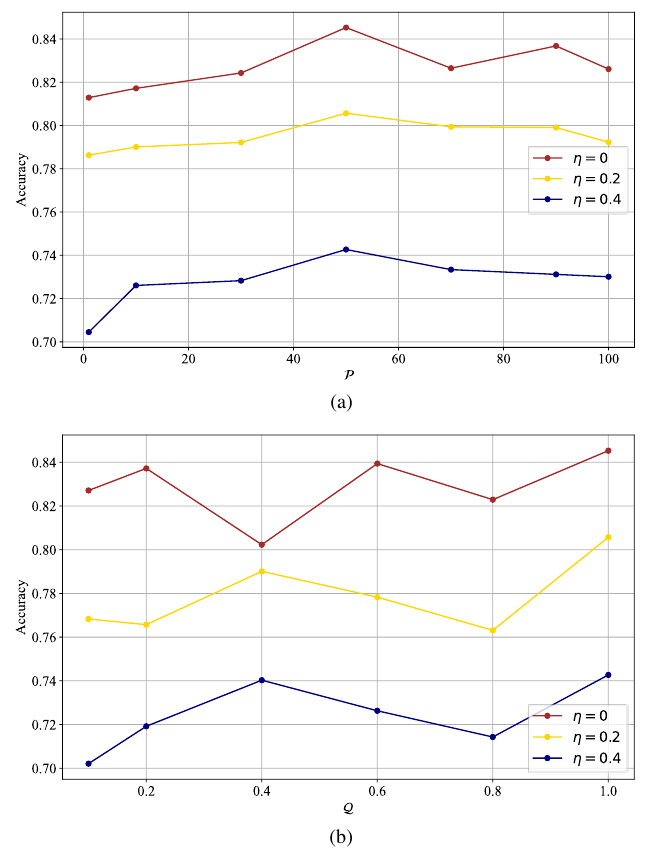}
 \caption{Parameter evaluation for $\lambda_{t} = \min(\mathcal{Q}, t/\mathcal{P})$ on BRCA. The figures show the impact of the annealing step (\(\mathcal{P}\) in panel (a)) and the converged value (\(\mathcal{Q}\) in panel (b)) on inference accuracy under different missing rates (\(\eta = 0, 0.2, 0.4\)). Both parameters affect the inference accuracy within a certain range, demonstrating their significance in the proposed method.}
\label{fig:para}
\end{figure}
We now provide detailed empirical results to investigate the above key questions, which can validate the effectiveness and trustworthiness of our model.

\textbf{Q1 Effectiveness (I).} 
As illustrated in Table~\ref{tab:acc}, we evaluate classification accuracy across multiple datasets at varying noise levels, \(\eta = 0\) to \(0.5\), simulating real-world scenarios where multi-view data is often randomly incomplete. 

At higher missing rate levels, particularly \(\eta = 0.5\), the proposed method consistently outperforms or competes favorably with existing approaches, including the UIMC framework which employs Dempster-Shafer combination rule as the way of fusion. Notably, the method's robustness to data corruption and loss becomes evident in its performance across various datasets. For instance, at \(\eta = 0.4\) and \(0.5\), the proposed method showcases superior accuracies, especially in datasets where robustness is crucial.

In the YaleB dataset at \(\eta = 0.4\), the method achieves an accuracy of 0.9265, significantly surpassing UIMC's 0.8440. Additionally, in the ROSMAP dataset at \(\eta = 0.5\), it records an accuracy of 0.7850, markedly better than UIMC's 0.7500. These examples highlight the proposed method's ability to maintain higher accuracy and demonstrate its effectiveness in handling severe data challenges compared to other methods. This performance underscores the method's potential as a robust and reliable tool for data analysis in adverse conditions, offering significant advantages in practical applications where data integrity cannot be guaranteed.

Moreover, consistent high performance across diverse datasets, from facial recognition to handwriting recognition underscores the method's versatility and generalizability. This adaptability is essential for practical applications, affirming the method’s potential as a robust and reliable tool for advanced data analysis in challenging conditions.

\textbf{Q2 Effectiveness (II).} 
In evaluating classification performance on large-scale multi-view datasets, specifically Caltech-101 and NUS-WIDE-OBJECT, the proposed method showcases robust effectiveness, particularly at higher missing rates (\(\eta = 0.4\) and \(0.5\)) as shown in Figure~\ref{fig:large_scale}. Unlike competing methods such as UIMC, DeepIMV, DCP, and CPM-Nets, which exhibit significant performance declines with increased missing rates, our method incorporates simple data imputation techniques before inference, yet allowing it to maintain superior accuracy. 

\begin{table*}[htbp]
\centering
\caption{Comparison in terms of classification accuracy with misaligned views (mean$\pm$std) with $\eta = [0, 0.1, 0.2, 0.3, 0.4, 0.5]$ on five datasets.}
\label{tab:misaligned}
\resizebox{1.0\textwidth}{!}{
\begin{tabular}{c|c|ccccccc}
\toprule
\multirow{2}{*}{Datasets}    & \multirow{2}{*}{Methods} & \multicolumn{6}{c}{Missing rates}                                                                         \\
                             &                          & $\eta= 0$               & $\eta=0.1$             & $\eta=0.2$             &$\eta=0.3$             & $\eta=0.4$             & $\eta=0.5$             \\
\midrule
\multirow{3}{*}{YaleB}       & UIMC                      & $0.9923\pm0.00$ & $0.9846\pm0.00$ & $0.9769\pm0.00$ & $0.9385\pm0.01$ & $0.8769\pm0.01$ & $0.7769\pm0.02$ \\
                             & Naive Fusion (Ours)                      & $\pmb{1.0000\pm0.00}$ & $0.9923\pm0.00$ & $0.9615\pm0.00$ & $0.9308\pm0.01$ & $0.8923\pm0.00$ & $0.8308\pm0.02$\\\cmidrule{2-8}
                             & Discount Fusion (Ours)                      & $\pmb{1.0000\pm0.00}$ & $\pmb{1.0000\pm0.00}$ & $\pmb{0.9846\pm0.00}$ & $\pmb{0.9692\pm0.01}$ & $\pmb{0.8923\pm0.00}$ & $\pmb{0.8538\pm0.01}$\\  
\midrule
\multirow{3}{*}{ROSMAP}      & UIMC                      & $0.7429\pm0.01$ & $0.7286\pm0.01$ & $0.7143\pm0.00$  & $0.6429\pm0.01$ & $0.6286\pm0.03$ &  $0.5857\pm0.02$ \\
                             & Naive Fusion (Ours)                      & $0.7635\pm0.01$ & $0.7429\pm0.01$ & $0.7143\pm0.01$ & $0.6829\pm0.01$ & $0.6429\pm0.01$ & $0.6171\pm0.02$\\\cmidrule{2-8}  
                             & Discount Fusion (Ours)                      & $\pmb{0.8019\pm0.01}$ & $\pmb{0.7922\pm0.01}$ & $\pmb{0.7714\pm0.01}$ & $\pmb{0.7549\pm0.01}$ & $\pmb{0.7286\pm0.01}$ & $\pmb{0.6928\pm0.01}$\\                           
\midrule
\multirow{3}{*}{Handwritten} & UIMC                      & $0.9800\pm0.00$          & $0.9800\pm0.00$         & $0.9750\pm0.01$          & $0.9650\pm0.02$ & $0.9600\pm0.02$ & $0.9350\pm0.02$ \\
                             & Naive Fusion (Ours)                     & $0.9825\pm0.00$ & $0.9775\pm0.01$ & $0.9650\pm0.01$ & $0.9475\pm0.02$ & $0.9150\pm0.01$ & $0.8800\pm0.03$\\\cmidrule{2-8}
                             & Discount Fusion (Ours)                      & $\pmb{0.9850\pm0.00}$ & $\pmb{0.9850\pm0.00}$ & $\pmb{0.9775\pm0.01}$ & $\pmb{0.9700\pm0.01}$ & $\pmb{0.9600\pm0.00}$ & $\pmb{0.9450\pm0.01}$\\                             
\midrule
\multirow{3}{*}{BRCA}        & UIMC                      & $0.7886\pm0.02$ & $0.7600\pm0.02$          & $0.7143\pm0.01$ & $0.6971\pm0.01$ & $0.6843\pm0.01$ & $0.6743\pm0.02$ \\
                             & Naive Fusion (Ours)                      & $0.7713\pm0.00$ & $0.7658\pm0.01$ & $0.7432\pm0.02$ & $0.7314\pm0.01$ & $0.7129\pm0.01$ & $0.6843\pm0.02$\\\cmidrule{2-8}
                             & Discount Fusion (Ours)                      & $\pmb{0.7929\pm0.01}$ & $\pmb{0.7743\pm0.00}$ & $\pmb{0.7650\pm0.01}$ & $\pmb{0.7429\pm0.01}$ & $\pmb{0.7143\pm0.01}$ & $\pmb{0.7092\pm0.01}$\\
\midrule
\multirow{3}{*}{Scene15}       & UIMC                      & $0.6957\pm0.01$ & $0.6934\pm0.01$ & $0.6678\pm0.01$ & $0.6253\pm0.01$ & $0.5908\pm0.01$ & $0.5426\pm0.01$\\
                             & Naive Fusion (Ours)                      & $0.7480\pm0.02$ & $0.6901\pm0.02$ & $0.6522\pm0.00$ & $0.6176\pm0.02$ & $0.5719\pm0.02$ & $0.5273\pm0.03$\\\cmidrule{2-8}
                             & Discount Fusion (Ours)                     & $\pmb{0.7865\pm0.01}$ & $\pmb{0.7476\pm0.01}$ & $\pmb{0.6895\pm0.02}$ & $\pmb{0.6577\pm0.01}$ & $\pmb{0.6265\pm0.01}$ & $\pmb{0.5873\pm0.01}$\\
\bottomrule
\end{tabular}}
\end{table*}

\textbf{Q3 Ablation study.} 
The Table~\ref{tab:noise} presents a detailed comparison of classification accuracy across five datasets, using methods based on the Dempster-Shafer theory, specifically UIMC, naive fusion, and discount fusion in terms of our proposed method. These datasets incorporate noisy views, potentially leading to conflicting evidence among views, as outlined in Section~\ref{sec:dataset}. These methods adopt distinct approaches to managing conflicts. While UIMC merges multi-view evidences directly, naive fusion averages evidences across the number of views \(V\). In contrast, discount Fusion implements a learnable dynamic discount factor that weights the evidence from each view prior to fusion.

As illustrated in Table~\ref{tab:noise}, discount fusion consistently outperforms both UIMC and naive fusion across all datasets and across increasing missing rates (\(\eta\)). This performance trend is particularly noticeable at higher missing rates, demonstrating that dynamic fusion in terms of learnable discount factors can help achieve superior capability in managing noisy and conflicting data through its dynamic adjustment of view-specific opinion's significance and consistency. For instance, in the YaleB dataset at \(\eta = 0.5\), discount fusion achieves an accuracy of 0.9000, substantially higher than 0.6769 by UIMC and 0.8538 by naive fusion. Similar patterns are observed in other datasets like ROSMAP and Handwritten, where discount fusion maintains the lead, notably achieving 0.9550 in Handwritten at \(\eta = 0.5\) compared to 0.8350 by UIMC. Even in datasets such as BRCA and Scene15, which exhibit overall lower accuracies due to increased challenges posed by missing views, discount fusion still presents the best performance figures, such as 0.7098 in BRCA and 0.5396 in Scene15 at \(\eta = 0.5\).

The Table~\ref{tab:misaligned} further underscores the robustness and efficacy of discount fusion when confronted with incomplete yet misaligned multi-view datasets. Overall, discount fusion significantly outperforms the other Dempster-Shafer based fusion framework. For example, in the YaleB dataset, the performance of discount fusion remains notably high even as the missing rate increases, achieving an accuracy of 0.8538 at $\eta=0.5$ compared to UIMC's 0.7769 and naive fusion's 0.8308. In the ROSMAP dataset, discount fusion achieves 0.6928 at $\eta=0.5$, significantly outperforming UIMC's 0.5857 and naive fusion's 0.6171. 

\textbf{Q4 Sensitivity of $\lambda_t$. }
Figure 3 demonstrates the sensitivity and necessity of hyperparameters \(\lambda_t\) in the proposed method, through a parameter evaluation focusing on the impact of the annealing step ($\mathcal{P}$) and the converged value (\(Q\)) on inference accuracy across varying missing rates (\(\eta = 0, 0.2, 0.4\)). The parameter \(\lambda_t\), defined as \(\min(P, t/Q)\), is crucial in modulating the KL divergence loss which in turn affects the model’s estimation of latent uncertainties, adjusting the assessment of uncertainty in erroneous evidences, and thereby progressively optimizing the quantification of latent uncertainties.

Figure~\ref{fig:para} (a) illustrates the influence of $\mathcal{P}$, the annealing step, on model accuracy. The curves demonstrate that for \(\eta = 0\), the accuracy remains relatively stable across different $\mathcal{P}$ values, maintaining high levels of accuracy. However, as missing rates increase (\(\eta = 0.2\) and \(0.4\)), the model exhibits higher sensitivity to $\mathcal{P}$, particularly noting a dip in accuracy at mid-range $\mathcal{P}$ values before recovering. This pattern suggests a critical dependence on the annealing step in environments with higher missing rates, indicating that a properly tuned $\mathcal{P}$ can effectively manage the increased uncertainty introduced by missing data.

Figure~\ref{fig:para} (b) shows the effect of the converged value (\(Q\)) on accuracy. Here, for each missing rate, the accuracy demonstrates a distinct peak at certain \(Q\) values before declining. Notably, the optimal \(Q\) value shifts slightly with the increase in \(\eta\), suggesting that the best \(Q\) is dependent on the degree of missing data in the views. This shift illustrates the adaptive nature of the model to different levels of data incompleteness, highlighting the hyperparameter's role in tailoring the uncertainty-driven inference process.

Overall, these results underscore the significant sensitivity and critical importance of the hyperparameters $\mathcal{P}$ and \(Q\) in the proposed method. Their careful calibration is shown to optimize inference results within certain parameter ranges, particularly under conditions of varied missing rates. The ability of these parameters to adaptively manage the latent uncertainties ensures that the model remains robust across different scenarios, thereby emphasizing the necessity of these hyperparameters in achieving optimal performance of the model. This not only confirms the method’s efficacy but also its adaptability, making it particularly valuable in practical applications where data incompleteness is prevalent.

\section{Conclusion}
\label{sec:conclusion}
This paper introduced a novel framework, Evidential Deep Partial Multi-View Classification, designed to tackle the significant challenges posed by incomplete multi-view data classification. By implementing K-means imputation for missing views and introducing a Conflict-Aware Evidential Fusion Network (CAEFN), the framework effectively addresses uncertainties and potential conflicts in imputed data, enhancing the reliability of inference outcomes. Comprehensive experiments conducted across various benchmark datasets demonstrate that our approach not only meets but frequently surpasses the performance of existing state-of-the-art methods, thereby validating the efficacy of our conflict-robust classification model in practical scenarios.

While the proposed method shows promising results, several areas could benefit from further investigation. The current approach primarily focuses on the fusion of evidential data without extensive exploration of deeper or more complex imputation strategies that could potentially capture finer details of missing data. Future work could explore integrating advanced neural architectures or unsupervised learning techniques to enrich the imputation process. Additionally, the sensitivity of the model to different types and degrees of data corruption suggests that further research into adaptive and context-aware fusion mechanisms could improve performance stability across even more diverse and challenging datasets. Finally, expanding the applicability of the model to other forms of incomplete data, such as time-series or irregularly sampled data, could significantly broaden its utility in real-world applications.

\printcredits

\bibliographystyle{model1-num-names}

\bibliography{cas-refs}

\end{document}